\documentclass[runningheads]{llncs}
\usepackage{graphicx}
\usepackage{amssymb}
\usepackage{threeparttable}
\usepackage[table]{xcolor}
\usepackage{stfloats}
\usepackage[font=small,skip=-1pt]{caption}
\definecolor{light-gray}{gray}{0.60}
\begin{document}
\makeatletter
\newcommand{\printfnsymbol}[1]{%
  \textsuperscript{\@fnsymbol{#1}}%
}
\makeatother

\title{Less is More: Simultaneous View Classification and Landmark Detection for Abdominal Ultrasound Images}
\titlerunning{Simultaneous View Classification and Landmark Detection}
\author{Zhoubing Xu\inst{1}\thanks{Equal Contribution} \and
Yuankai Huo\inst{2}\printfnsymbol{1} \and
JinHyeong Park\inst{1} \and 
Bennett Landman\inst{2} \and 
Andy Milkowski\inst{3} \and 
Sasa Grbic\inst{1} \and 
Shaohua Zhou\inst{4}\thanks{This work was done when Zhou was with Siemens Healthineers}}
\authorrunning{Z. Xu and Y. Huo et al.}
%
\institute{Siemens Healthineers, Medical Imaging Technologies, Princeton, NJ, USA \and
Vanderbilt University, Electrical Engineering, Nashville, TN, USA \and
Siemens Healthineers, Ultrasound, Issaquah, WA, USA \and 
Institute of Computing Technology, Chinese Academy of Sciences, Beijing, China}
\maketitle              
\begin{abstract}
An abdominal ultrasound examination, which is the most common ultrasound examination, requires substantial manual efforts to acquire standard abdominal organ views, annotate the views in texts, and record clinically relevant organ measurements. Hence, automatic view classification and landmark detection of the organs can be instrumental to streamline the examination workflow. However, this is a challenging problem given not only the inherent difficulties from the ultrasound modality, e.g., low contrast and large variations, but also the heterogeneity across tasks, i.e., one classification task for all views, and then one landmark detection task for each relevant view. While convolutional neural networks (CNN) have demonstrated more promising outcomes on ultrasound image analytics than traditional machine learning approaches, it becomes impractical to deploy multiple networks (one for each task) due to the limited computational and memory resources on most existing ultrasound scanners. To overcome such limits, we propose a multi-task learning framework to handle all the tasks by a single network. This network is integrated to perform view classification and landmark detection simultaneously; it is also equipped with global convolutional kernels, coordinate constraints, and a conditional adversarial module to leverage the performances. In an experimental study based on 187,219 ultrasound images, with the proposed simplified approach we achieve (1) view classification accuracy better than the agreement between two clinical experts and (2) landmark-based measurement errors on par with inter-user variability. The multi-task approach also benefits from sharing the feature extraction during the training process across all tasks and, as a result, outperforms the approaches that address each task individually.
\end{abstract}
\section{Introduction}
Ultrasound scanning is widely used for safe and non-invasive clinical diagnostics. Given a large population with gastrointestinal diseases (60-70 million in USA \cite{peery2012burden}), the abdomen is one of the most commonly screened body parts under ultrasound examinations. During an abdominal examination session, a sonographer needs to navigate and acquire a series of standard views of abdominal organs, annotate the view information in texts, adjust the caliper to desirable locations through a track ball, and record measurements for each clinically relevant organ. The substantial manual interactions not only become burdensome for the user, but also decrease the workflow efficiency.

Automatic view classification and landmark detection of the abdominal organs on ultrasound images can be instrumental to streamline the examination workflow. However, it is very challenging to accomplish the full automation from two perspectives. First, analytics of the ultrasound modality are inherently difficult due to the low contrast and large variations throughout ultrasound images, which are sometimes confusing even to experienced ultrasound readers. Second, the associated tasks are typically handled individually, i.e., one classification task for all views, and then one landmark detection task for each relevant view, due to their heterogeneities between each other; this is very hard to fulfill on most existing ultrasound scanners, restricted by limited computational and memory resources.
\begin{figure}[t]
\begin{center}
\includegraphics[scale=0.5]{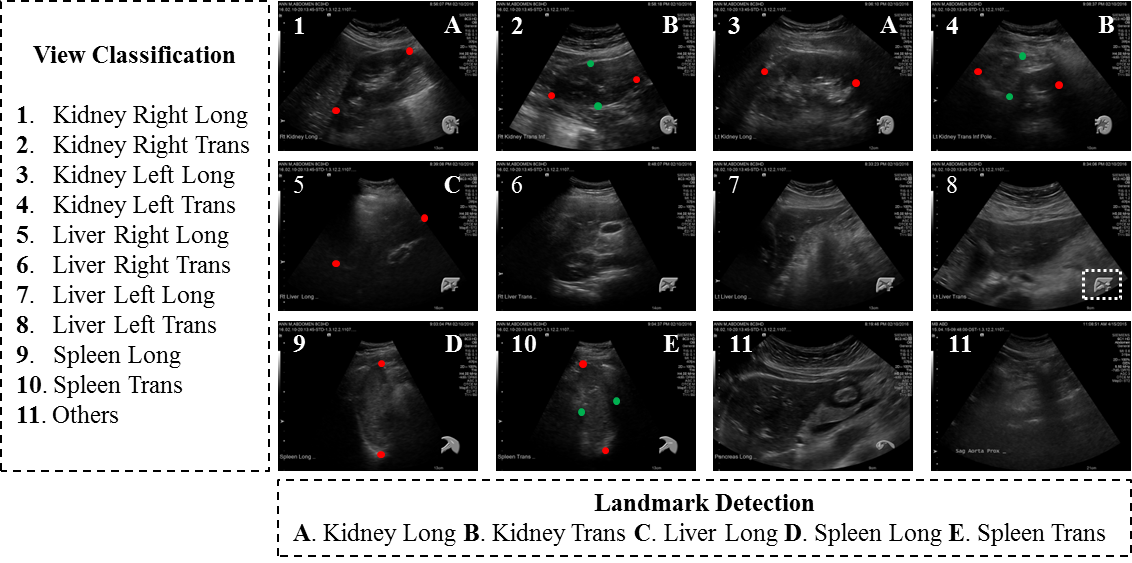}
\caption{An overview of the tasks for abdominal ultrasound analytics. In each image, the upper left corner indicates its view type. If present, the upper right corner indicates the associated landmark detection task, and the pairs of long- and short-axis landmarks are colored in red and green, respectively. An icon is circled on one image; such icons are masked out when training the view classification.} \label{fig1}
\end{center}
\end{figure}
Convolutional neural networks (CNN) have demonstrated superior performance to traditional machine learning methods given large-scale datasets in many medical imaging applications \cite{ronneberger2015u}. they provide a favorable option to address ultrasound problems. Based on CNN, multi-task learning (MTL) has been investigated to improve outcomes for each single task with the assumption that common hidden representations are shared among multiple tasks. In MTL, a single neural network is used instead of one network per task so that the requirement for computational and memory resources is substantially reduced. Therefore, we pursue a highly integrated MTL framework to perform simultaneous view classification and landmark detection automatically to increase the efficiency of abdominal ultrasound examination workflow.

Current researches on MTL are quite diversified given the task varieties. Kokkinos et al. \cite{kokkinos2017ubernet} presented a unified framework to accomplish seven vision tasks on a single image. The tasks are highly correlated, but differ from each other by focusing on different levels of image details. Given these tasks, it turns out helpful to extract comprehensive image features by sharing every level of convolutional layers, and then branching out for task-specific losses at each level and each resolution. Ranjan et al. \cite{ranjan2017all} introduced another unified framework for detection of face attributes. With empirical knowledge of what level of features can best represent each face attribute, the branching location for each task is customized in this all-in-one network to maximize the synergy across all the tasks. Xue et al. \cite{xue2017full} demonstrated full quantification of all size-related measurements of left ventricle in cardiac MR sequences. They also incorporated the phase information as an additional task, through which they regularized the intra- and inter-task relatedness along sequences. While these studies achieved successes in their applications, they cannot be easily adapted to our problem. They are mostly designed for estimating variable attributes of a single object or on a single image, while our scenario is to predict multiple attributes (view and landmarks) on different objects (e.g., liver, spleen, kidney) that are not in the same image for most cases. Moeskops et al. \cite{moeskops2016deep} used a single network architecture to perform diverse segmentation tasks (brain MR, breast MR, and cardiac CTA) without task-specific training; this is similar to our problem for handling different organs, but simpler with only segmentation tasks.

In this study, we propose an end-to-end MTL network architecture tailored for abdominal ultrasound analytics to synergize the extraordinary heterogeneous tasks. To the best of our knowledge, we are the first to present an integrated system with fully automated functionalities for the abdominal ultrasound examination. 

\section{Methods and Results}
\subsection{Data and Task Definitions}
During a typical abdominal ultrasound session, ten standard views are of major interests, including five structures from two orientations, i.e., liver right lobe, liver left lobe, right kidney, left kidney, and spleen, from longitudinal and transverse directions. On certain views, two longitudinal (“Long”) or four transverse (“Trans”) landmarks are placed for measurements, where each measurement is derived from a pair of landmarks along the long- or short- axis. These processes are summarized as an 11-view classification task (with an addition of the “others” view) and 5 landmark detection tasks with a total of 14 landmarks (Fig.~\ref{fig1}). Note that we combine the left and right views of kidneys for landmark detection.
\begin{figure}[t]
\begin{center}
\includegraphics[scale=0.5]{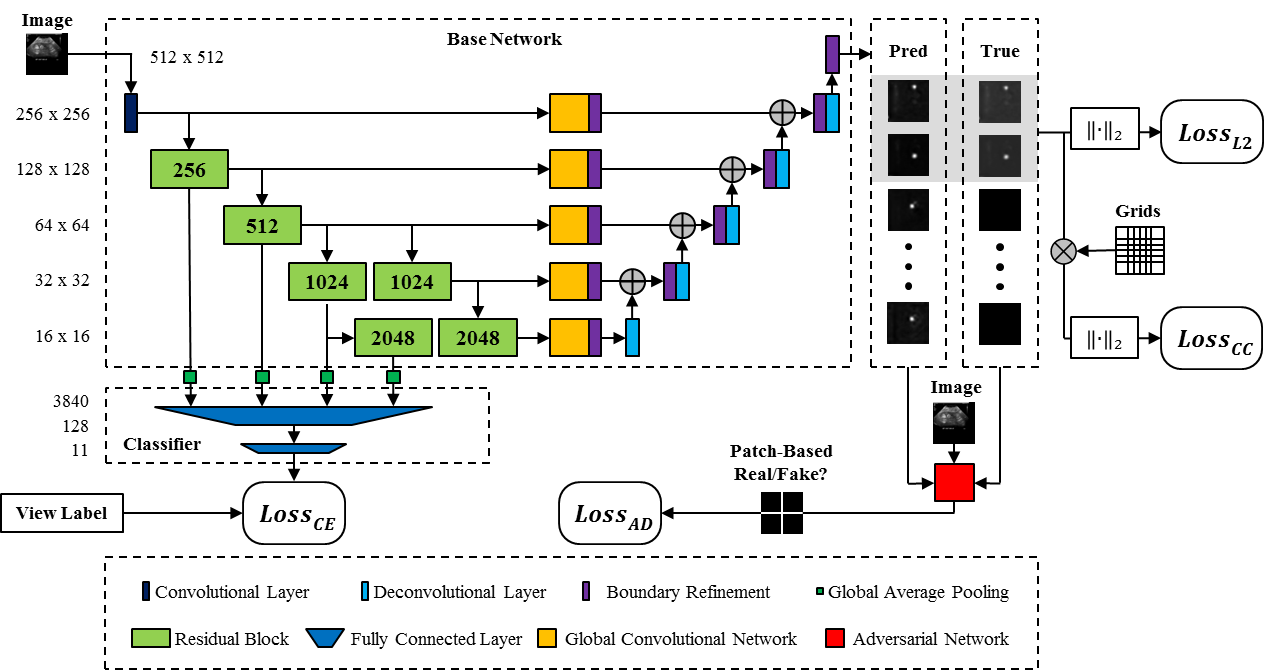}
\end{center}
\caption{An illustration of the proposed MTL Framework.} \label{fig2}
\end{figure}
A total of 187,219 ultrasound images from 706 patients were collected. Some images are acquired in sequence, while others are as single frames. The view information was manually assigned to each image as a routine during acquisition, where a representative icon was placed at the corner of the image. Landmark annotations were performed by experienced ultrasound readers on 7,012 images (1921, 1309, 1873, 1711, and 198 for the five landmark detection tasks listed in Fig.~\ref{fig1}), and further verified by the ultrasound specialist. Duplicated annotations were performed by a second ultrasound expert for inter-user variabilities on 1999 images for views and 30 \- 50 images for each pair of landmarks. The datasets were separated on the patient level into training and testing sets by an 80\%/20\% split for all tasks. Any patients with data included in the training set were excluded from the testing set. During preprocessing, the images were resampled into 0.5mm isotropic resolution, and zero-padded to 512 \(\times\) 512. For each image used for training the view classification, a mask was applied to block view-informative icons and texts. The landmark annotations were converted into distance-based Gaussian heat maps centered to the landmark locations.
\subsection{MTL Framework}
In the proposed MTL network architecture (Fig.~\ref{fig2}), we construct an encoder that follows ResNet50 \cite{he2016deep}.  The first convolutional layer is modified to take a single channel input. While the first two residual blocks are shared across all tasks for low level feature extraction, two copies of the 3rd and 4th residual blocks are used, one for view classification, and the other for landmark detection.

\textbf{Cross Entropy Loss for View Classification:} For view classification, each residual block of its encoder is connected with a global average pooling (GAP) layer, and a feature vector is composed by concatenating the pooled features from multiple levels; two fully connected layers are used as the classifier based on the pooled features. A traditional cross entropy loss is defined as 
\begin{equation}
Loss_{CE} = -\sum_c y_c \log p_c
\end{equation}
\noindent where \(y_c\) represents the binary true value, and \(p_c\) indicates the probability of an image being view class \(c\), respectively.

\textbf{Regression Loss for Landmark Detection:} For landmark detection, we form a single decoder shared across the landmarks from all views instead of one branch per view, and the decoder follows the skip-connection style in Fully Convolutional Network (FCN) \cite{long2015fully}, where the output channel of each level is kept the same as the total number of landmarks, i.e., \(N_L\). On each level of skip connection between encoder and decoder, we append Global Convolutional Network (GCN) \cite{peng2017large} and boundary refinement modules to capture larger receptive fields. Consider \(L = \{0,1,…,N_L-1\}\) as the complete set of landmarks, let \(\hat{H} ̂\in \mathbb{R}^{N_L \times N_I}\) and \(H ̂\in \mathbb{R}^{N_L \times N_I}\) be the Gaussian heat maps for the prediction and truth, respectively, including all \(N_L\) channels of images (indexed with \(l\)), each with \(N_I\) pixels (indexed with \(i\)), an L2-norm is computed only on a selective subset \(L^{'} \subset L\) with \(N_{L^{'}}\) landmarks associated to each image so that only relevant information gets back-propagated
\begin{equation}
Loss_{L2} = \frac{1}{N_I N_{L^{'}}}  \sum_i \sum_{l \in L^{'}} \left(\hat{H}_{li} - H_{li}\right)^2
\end{equation}

\textbf{Landmark Location Error:} A coordinate-based constraint is applied to regularize the heat map activation. Consider the image grid coordinates as \(S \in \mathbb{R}^{N_I \times 1}\) and \(T \in \mathbb{R}^{N_I \times 1}\) for the two dimensions, the predicted landmark location \(\left(\hat{s}_l, \hat{t}_l\right)\) can be derived as a weighted average, i.e., \(\hat{s} = \frac{\sum_{i \in \Omega_l} H_{li} S_i}{\sum_{i \in \Omega_l} H_{li}}\), \(\hat{t} = \frac{\sum_{i \in \Omega_l} H_{li} T_i}{\sum_{i \in \Omega_l} H_{li}}\), where \(\Omega_l\) indicates the area above a threshold \(k\), i.e., \(H_l > k\). Unlike identifying the point with maximum value, this weighted averaging is a differentiable process to maintain the end-to-end training workflow. Then Euclidean distance error is computed against the true landmark location \(\left(s_l, t_l\right)\), 
\begin{equation}
Loss_{CC} = \frac{1}{N_{L^{'}}} \sum_{l \in L^{'}} \sqrt{\left(s_l - \hat{s}_l\right)^2 + \left(t_l - \hat{t}_l\right)^2}
\end{equation}

\textbf{Adversarial Loss:} Following PatchGAN \cite{isola2017image}, an adversarial network \(\textbf{D}: \mathbb{R}^{N_I \times \left(N_L + 1\right)} \mapsto \mathbb{R}^{N_M \times 1}\) is defined; it takes both the input image and the output prediction from the base network to identity the real and fake outputs on the basis of \(N_M\) patches. From the perspective of the base network, it regularizes the output with binary cross entropy
\begin{equation}
Loss_{AD} = \frac{1}{N_{M}} \sum_m [-\sum_{i \in m} \left(y_i \log{p_i} + \left(1 - y_i\right) \log{1 - p_i}\right)]
\end{equation}
\noindent where \(m\) indicates a single patch, \(y\) is the real true/false, and \(p\) is the probability of true/false prediction. This effectively enforces the output heat maps to follow a reasonable landmark distribution. 
\begin{figure}[t]
\begin{center}
\includegraphics[scale=0.5]{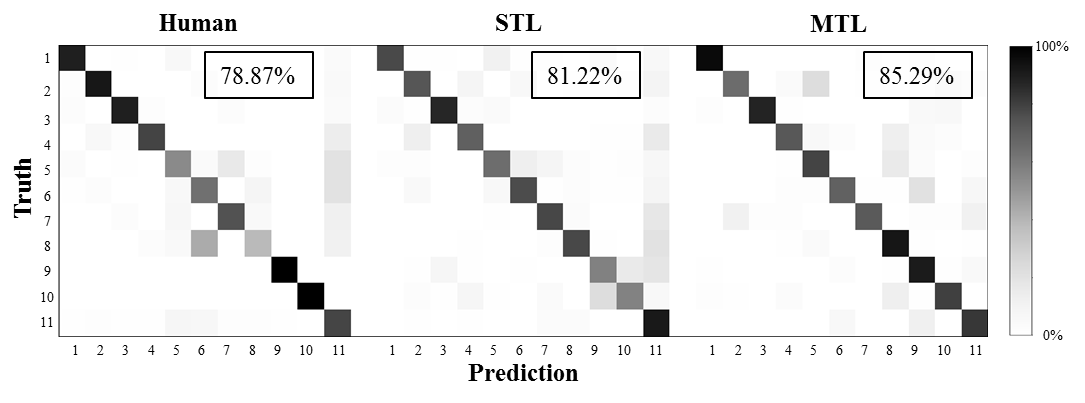}
\end{center}
\caption{Confusion matrices for the view classification task of STL, MTL, and between humans. The numbers on x and y axes follow the view definitions in Fig.~\ref{fig1}. The overall classification accuracy is overlaid for each approach. The diagonal entries indicate correct classifications.} \label{fig3}
\end{figure}

\textbf{Implementation:} The landmark detection tasks are trained first with a batch size of 4 for 30 epochs, where each batch can include a mixture of different organs. The base network is optimized by stochastic gradient decent (SGD) with a learning rate (LR) of 1E-6. The adversarial network uses the Adam optimizer with a LR of 2E-4. We take \(k=0.75\) as the threshold for the coordinate derivation. The regression, location, and adversarial losses are equally weighted as an empirical configuration. The view classification task is trained afterwards with a batch size of 8 for 5 epochs, and optimized by Adam with a LR of 5E-4, while the shared parameters pre-trained with landmark detection tasks are kept locked. During testing, the image can be forwarded though the network for all tasks by one shot. The experiments are performed on a Linux workstation equipped with an Intel 3.50 GHz CPU and a 12GB NVidia Titan X GPU using the PyTorch framework. Single-task learning (STL) approaches and ablation studies are performed for comparison using the same configuration as the proposed method as much as possible. The STL view classification takes the default ResNet50 pre-trained from ImageNet \cite{russakovsky2015imagenet} with the data pre-processed accordingly. For evaluation based on clinical standards, we use the classification accuracy for view classification, and the absolute differences of the long- and short-axis measurements for landmark detection. 
\subsection{Results}
\begin{figure}[t]
\begin{center}
\includegraphics[width=0.60\textwidth]{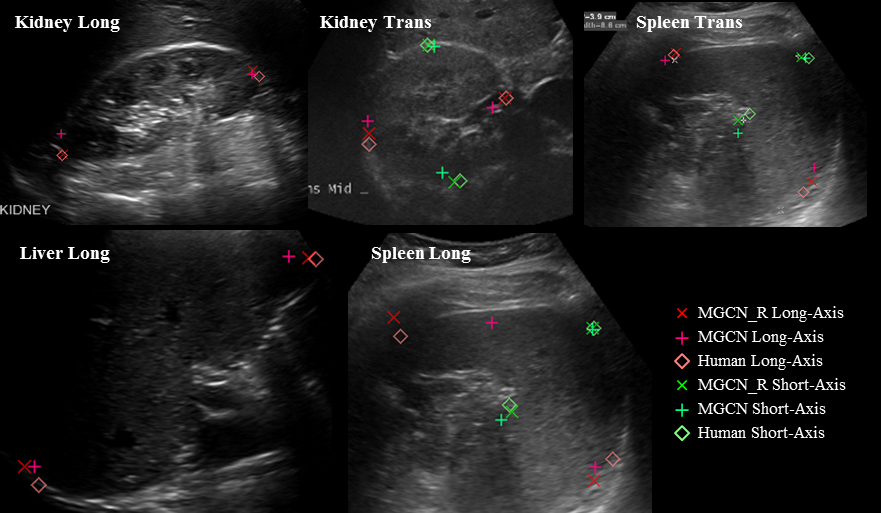}
\end{center}
\caption{Qualitative comparison of landmark detection results with and without regularization for the proposed approach. Images are zoomed into region of interest for better visualization} \label{fig4}
\end{figure}

For view classification, we achieve a 4.07\% improvement compared to STL, and we also outperform the second human expert, especially for distinguishing the non-others classes (Fig.~\ref{fig3}). Please note that we use the annotations from one expert with more ultrasound experience as the ground truth for training, and those from the other less experienced expert for reference. For landmark detection, we reduce each landmark-based measurement error by a large margin compared to every other benchmark approaches (Table ~\ref{tab1}). For most measurements, we also achieve errors below 1.5 times of the inter-user variability. The two measurements for Spleen Trans are slightly worse due to lack of samples (< 3\% of total). Going directly from STL to MTL (SFCN\(\rightarrow\)MFCN) provides implicit data augmentation for tasks with limited data, while the accuracies on other tasks seem to be compromised. Using GCN, the variabilities from multiple tasks are better captured, which leads to improved results (MFCN\(\rightarrow\)MGCN). With the coordinate-based constraint and patch-based adversarial regularization on the outputs, the outliers of landmark detection can be substantially reduced (MGCN\(\rightarrow\)MGCN\_R), and thus boosts the performance (Fig.~\ref{fig4}). With the GPU implementation, the average time consumption is 90ms to load and pre-process the data, classify the view, detect landmarks, and derive measurements. The model parameters to handle all the tasks are reduced from 539MB to 100MB.
\section{Discussion}
In this paper, we propose an end-to-end MTL framework that enables efficient and accurate view classification and landmark detection for abdominal ultrasound examination. The main novelty lies in (1) the integration of the heterogeneous tasks into a single network, (2) the design of two regularization criteria to improve MTL performance, and (3) the first systematic design of a streamlined end-to-end workflow for abdominal ultrasound examination. 
\setlength\belowcaptionskip{0ex}
\begin{threeparttable}[hb]
\caption{Quantitative comparison for landmark-based measurements in mm.}\label{tab1}
\begin{tabular}{|p{0.13\textwidth} | p{0.12\textwidth} | p{0.12\textwidth} | p{0.12\textwidth} | p{0.12\textwidth} | p{0.12\textwidth} | p{0.12\textwidth} | p{0.12\textwidth} |}
\hline
\ & KL\_LA &  KT\_LA & KT\_SA & LL\_LA & SL\_LA & ST\_LA & ST\_SA\\
\hline
\rowcolor{light-gray}
Human & 4.500 & 5.431 & 4.283 & 5.687 & 6.104 & 4.578 & 4.543\\
\hline
PBT \cite{tu2005probabilistic} & 11.036 & 9.147 & 8.393 & 11.083 & 7.289 & 9.359 & 12.308\\
\hline
SFCN & 7.044 & 7.332 & 5.189 & 10.731 & 8.693 & 91.309 & 43.773\\
\hline
MFCN & 10.582 & 16.508 & 15.942 & 17.561 & 8.856 & 49.887 & 29.167\\
\hline
MGCN & 10.628 & 5.535 & 5.348 & 9.46 & 7.718 & 12.284 & 19.79\\
\hline
MGCN\_R & \textbf{4.278} & \textbf{4.426} & \textbf{3.437} & \textbf{6.989} & \textbf{3.610} & \textbf{7.923} & \textbf{7.224}\\
\hline
\end{tabular}
\begin{tablenotes}
\item [] Note that LA and SA represent for long- and short-axis measurements. KL, KT, LL, SL, and ST stand for Kidney Long, Kidney Trans, Liver Long, Spleen Long, and Spleen Trans, respectively. For the methods, the prefix S and M represent single-task and multi-task, respectively, while FCN and GCN are both based on ResNet50 except that the later embeds large kernels and boundary refinement in skip connection. MGCN\_R is the proposed method that includes two additional regularization modules. PBT is a traditional machine learning approach. The human statistics are computed on a subset of images for reference.\\
\end{tablenotes}
\end{threeparttable}

It is critical to determine where to share and diversify the heterogeneous tasks integrated in a single network. The landmark detection of different views can be considered a MTL problem by itself. Formulating the network with multiple decoders seems straightforward, but not favorable in this study, we observe overwhelming outliers with such network design in our preliminary tests. Our experiment demonstrates that sharing all landmark tasks with one decoder together with a selective scheme for back-propagation provides an effective training platform for the heterogeneous tasks of landmark detection. As the landmarks are defined similarly on the long- and short- axis endpoints of organs even though distributed on different views, it makes sense to share how the output heat maps are reconstructed from the extracted features with a single decoder. The share-it-all design also enables mixed organ types in one mini-batch to get back-propagated together, and thus augments the data implicitly; this is not simple with the multi-decoder design. View classification, on the other hand, is less compatible with the landmark detection tasks; there are also additional views involved. However, it still benefits from sharing the low level features. The classification accuracy gets improved by combining these shared low level features with the high level features learned by the view classification individually.

\noindent \textbf{Disclaimer:} This feature is based on research, and is not commercially available. Due to regulatory reasons its future availability cannot be guaranteed.

%
\bibliographystyle{splncs}
\bibliography{refs}
\end{document}